\documentclass{article}

\usepackage{microtype}
\usepackage{graphicx}
\usepackage{subcaption}
\usepackage{booktabs} % for professional tables

\usepackage{hyperref}

\usepackage[accepted]{icml2026}

\usepackage{amsmath}
\usepackage{amssymb}
\usepackage{mathtools}
\usepackage{amsthm}

\usepackage{cuted}      % 提供 strip 环境
\usepackage{caption}    % 提供 \captionof 命令
\usepackage{booktabs}
\usepackage{multirow}
\usepackage{colortbl}
\definecolor{C1}{HTML}{660874}
\usepackage{threeparttable}
\usepackage{wrapfig} % for figure

\usepackage[capitalize,noabbrev]{cleveref}

\theoremstyle{plain}

\theoremstyle{definition}

\theoremstyle{remark}

\usepackage[textsize=tiny]{todonotes}

\icmltitlerunning{Principled RL for Flow Matching Emerges from the Chunk-level Policy Optimization}

\begin{document}

\twocolumn[{
  \icmltitle{Principled RL for Flow Matching Emerges \\
from the Chunk-level Policy Optimization}

  % It is OKAY to include author information, even for blind submissions: the
  % style file will automatically remove it for you unless you've provided
  % the [accepted] option to the icml2026 package.

  % List of affiliations: The first argument should be a (short) identifier you
  % will use later to specify author affiliations Academic affiliations
  % should list Department, University, City, Region, Country Industry
  % affiliations should list Company, City, Region, Country

  % You can specify symbols, otherwise they are numbered in order. Ideally, you
  % should not use this facility. Affiliations will be numbered in order of
  % appearance and this is the preferred way.
  \icmlsetsymbol{equal}{*}
  \icmlsetsymbol{lead}{†}
  \icmlsetsymbol{work}{‡}

  \begin{icmlauthorlist}
    \icmlauthor{Yifu Luo}{equal,work,yyy}
    \icmlauthor{Haoyuan Sun}{equal,yyy}
    \icmlauthor{Xinhao Hu}{equal,yyy}
    \icmlauthor{Penghui Du}{equal,comp}
    \icmlauthor{Keyu Fan}{yyy}
    \icmlauthor{Bo Li}{lead,comp}
    \icmlauthor{Sinan Du}{yyy}
    \icmlauthor{Xu Wan}{xxx}
    \icmlauthor{Zhiyu Chen}{yyy}
    \icmlauthor{Bo Xia}{yyy}
    \icmlauthor{Yongzhe Chang}{yyy}
    \icmlauthor{Changqian Yu}{comp}
    \icmlauthor{Kun Gai}{comp}
    \icmlauthor{Tiantian Zhang}{yyy}
    %\icmlauthor{}{sch}
    % \icmlauthor{Firstname8 Lastname8}{sch}
    \icmlauthor{Xueqian Wang}{yyy}
    %\icmlauthor{}{sch}
    %\icmlauthor{}{sch}
  \end{icmlauthorlist}

  \icmlaffiliation{yyy}{Tsinghua University}
  \icmlaffiliation{xxx}{Zhejiang University}
  \icmlaffiliation{comp}{Kolors Team, Kuaishou Technology}
%   \icmlaffiliation{sch}{School of ZZZ, Institute of WWW, Location, Country}

  \icmlcorrespondingauthor{Changqian Yu}{yuchangqian@kuaishou.com}
  \icmlcorrespondingauthor{Tiantian Zhang}{zhang.tt@sz.tsinghua.edu.cn}

  % You may provide any keywords that you find helpful for describing your
  % paper; these are used to populate the "keywords" metadata in the PDF but
  % will not be shown in the document
  \icmlkeywords{Machine Learning, ICML}

  \vskip 0.1in
  \centering
  \includegraphics[width=0.78\textwidth]{figures/quality_main.jpg}
  \vskip -0.15in
  \captionof{figure}{Compared to GRPO, Group Chunking Policy Optimization (GCPO) yields a significant improvement in image quality, particularly regarding structure, lighting, and fine-grained details, demonstrating the superiority of chunk-level policy optimization. Notably, Notably, the contrast between columns $3$ and $4$  underscores the importance of chunking with the inherent temporal dynamics of flow matching, which substantially elevates aesthetic style and structure of images. Additional details are provided in \Cref{app_fig_detail}.}
  \label{figure_quality_main}}
  \vskip 0.22in
]
% this must go after the closing bracket ] following \twocolumn[ ...

% This command actually creates the footnote in the first column listing the
% affiliations and the copyright notice. The command takes one argument, which
% is text to display at the start of the footnote. The \icmlEqualContribution
% command is standard text for equal contribution. Remove it (just {}) if you
% do not need this facility.

% Use ONE of the following lines. DO NOT remove the command.
% If you have no special notice, KEEP empty braces:
% \printAffiliationsAndNotice{}  % no special notice (required even if empty)
% Or, if applicable, use the standard equal contribution text:
\printAffiliationsAndNotice{\icmlEqualContribution \textsuperscript{†}Project lead. \textsuperscript{‡}Work done during internship in Kolors Team, Kuaishou Technology.}

\begin{abstract}
  Recent Progress in post-training flow matching for text-to-image (T2I) generation with Group Relative Policy Optimization (GRPO) has demonstrated strong potential. However, it is hindered by a critical limitation: inaccurate advantage attribution. In this work, we argue that aggregating consecutive steps into a coherent `chunk' and shifting the policy optimization paradigm from GRPO's step level to the chunk level can effectively mitigate the negative impact of this issue. Building on this insight, we propose Group Chunking Policy Optimization (GCPO), the first chunk-level reinforcement learning approach for post-training flow matching. Extensive experiments demonstrate that GCPO achieves superior performance on both standard T2I benchmarks and preference alignment, with up to $43\%$ relative gains over GRPO, highlighting the promise of chunk-level policy optimization. The code is available on \url{https://github.com/xingzhejun/GCPO}.
\end{abstract}

\section{Introduction}

Reinforcement learning (RL) \cite{sutton1998reinforcement, schulman2017proximal} has recently achieved remarkable success beyond traditional domains, most notably in the post-training of Large Language Models (LLMs) \cite{jaech2024openai, guo2025deepseek, wang2026language, wang2026lifelong}. Inspired by these advances, recent works \cite{xue2025dancegrpo, liu2026flow, wang2025coefficients, luo2026reinforcement} have explored adapting RL to text-to-image (T2I) generation. In this context, Group Relative Policy Optimization (GRPO) \cite{shao2024deepseekmath, guo2025deepseek} has emerged as a promising approach for post-training flow matching \cite{lipman2022flow, liuflow, esser2024scaling} for T2I generation. Typically, GRPO methods generate a group of images from the same prompt, obtain final rewards using reward models, convert them into group relative advantages, and assign these advantages uniformly across all steps in flow matching generations for policy optimization.

\begin{figure}[ht]
  % \vskip 0.2in
  \centering
    \includegraphics[width=1.0\linewidth]{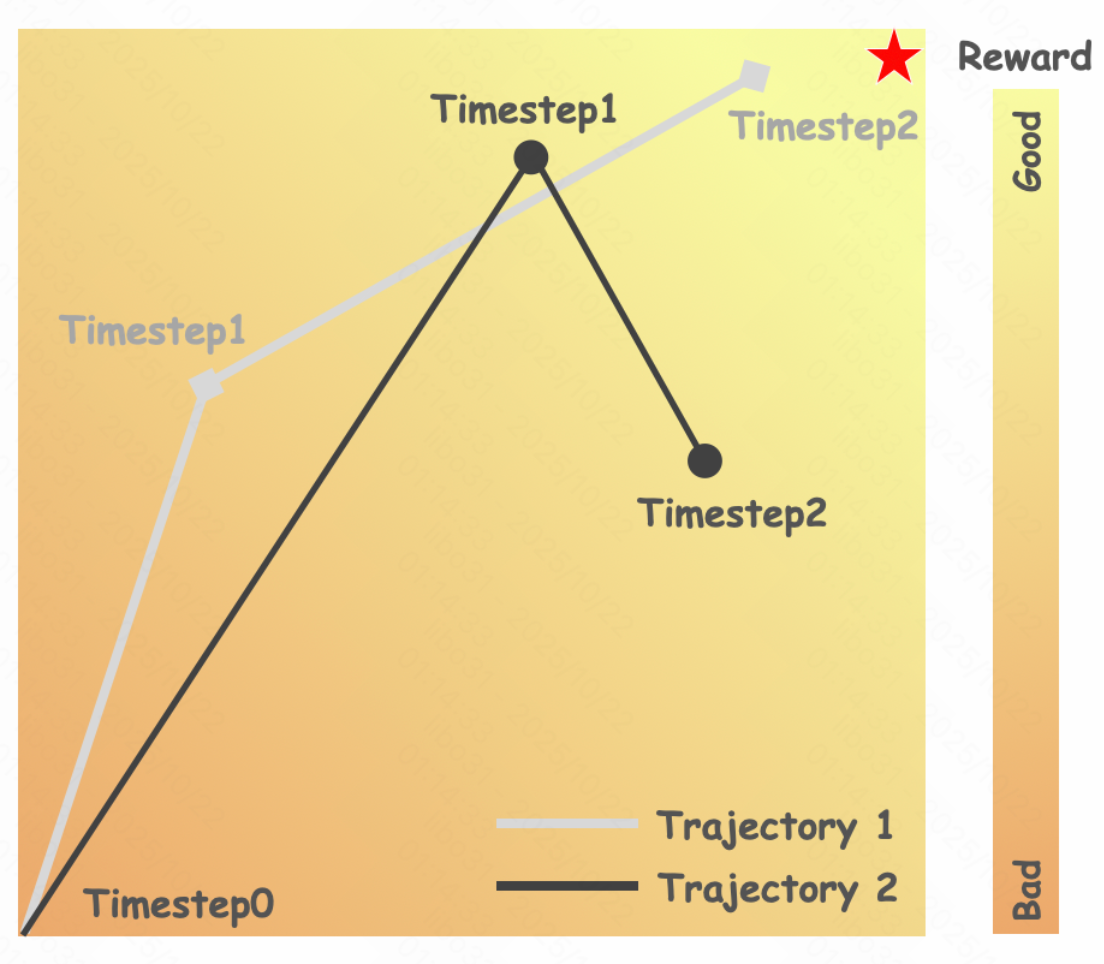}
    \caption{
      An illustration of the inaccurate advantage attribution. Although $\text{Trajectory}_1$ yields a better final reward, $\text{Trajectory}_2$ demonstrates a superior policy at step $t=1$. This conflicts with GRPO, which uniformly assigns the final advantage across all steps.
    }
    \label{figure_motivation} 
    % \vskip -0.02in
\end{figure}

Despite its effectiveness, GRPO methods suffer from a critical limitation: inaccurate advantage attribution. The equal attribution of final reward to all steps rests on a strong assumption: A superior final outcome implies a superior policy at every step. However, this assumption is not always guaranteed. Consider two generation trajectories from the same prompt in \Cref{figure_motivation}, each consisting of three steps. Although $\text{Trajectory}_1$ yields a better final reward, $\text{Trajectory}_2$ may have a superior policy at an intermediate step $t=1$. In such cases, GRPO provides erroneous optimization signals, resulting in training instability. To quantify it more generally, we utilized a step-aware preference model from \cite{liang2025aesthetic} to determine how often this occurs. As shown in \Cref{table_motivation}, approximately half of the steps are affected by this inaccurate attribution, further demonstrating its severity.

\begin{table}[t]
  \caption{Frequency of inaccurate advantage attribution. Approximately half of the steps are affected by it. Statistics are reported over $400$ prompts from HPDv2.1 \cite{wu2023human}. Further details are provided in \Cref{Details for the Frequency of Inaccurate Advantage Attribution}.}
  \label{table_motivation}
  \begin{center}
    \begin{small}
      \begin{sc}
        \begin{tabular}{l|lcc}
        \toprule
           \multicolumn{1}{l}{} & & \multicolumn{2}{c}{Final Reward} \\ 
\cmidrule(lr){3-4} % 局部横线，(lr)让线左右缩进一点，更好看
  \multicolumn{1}{l}{} & & better & worse \\ 
\midrule
\multirow{2}{1.6cm}{\centering Step-Advantage} & better & $63\%$ & $37\%$ \\ 
 & worse & $44\%$ & $56\%$ \\ 
\bottomrule
        \end{tabular}
      \end{sc}
    \end{small}
  \end{center}
 \vskip -0.1in
\end{table}

A straightforward solution to this issue would be to employ a precise, process (step-aware) reward model, which is capable of evaluating not only final clear images but also noisy intermediate images during the flow matching generation process. However, training such models requires vast amounts of preference data for noisy images, which is currently unavailable. While some approaches \cite{liang2025aesthetic, liao2025step} attempt to obtain process rewards using estimations like one-step diffusion, their effectiveness is limited due to estimation bias and the lack of ground-truth data \cite{zhang2025lessons}. Our comparison results in \Cref{Details for the Process Reward Comparison} further corroborate this.

In this work, instead of introducing process rewards, we mitigate the negative impact of this issue from a different perspective. We draw inspiration from action chunking \cite{zhao2023learning, li2025reinforcement} in robotics, which predicts a sequence of consecutive actions jointly as a single action rather than treating them independently. In a similar spirit, we propose to aggregate consecutive flow matching steps into a single ``action chunk", and optimize the policy at the chunk level rather than the step level. This effectively smooths out the gradient fluctuations caused by inaccurate advantage attribution, thereby stabilizing training. Furthermore, we leverage the distinct temporal dynamics \cite{wimbauer2024cache, liu2025timestep} in flow matching to naturally guide the chunking process.

\begin{figure*}
\centering
\includegraphics[width=1.0\textwidth]{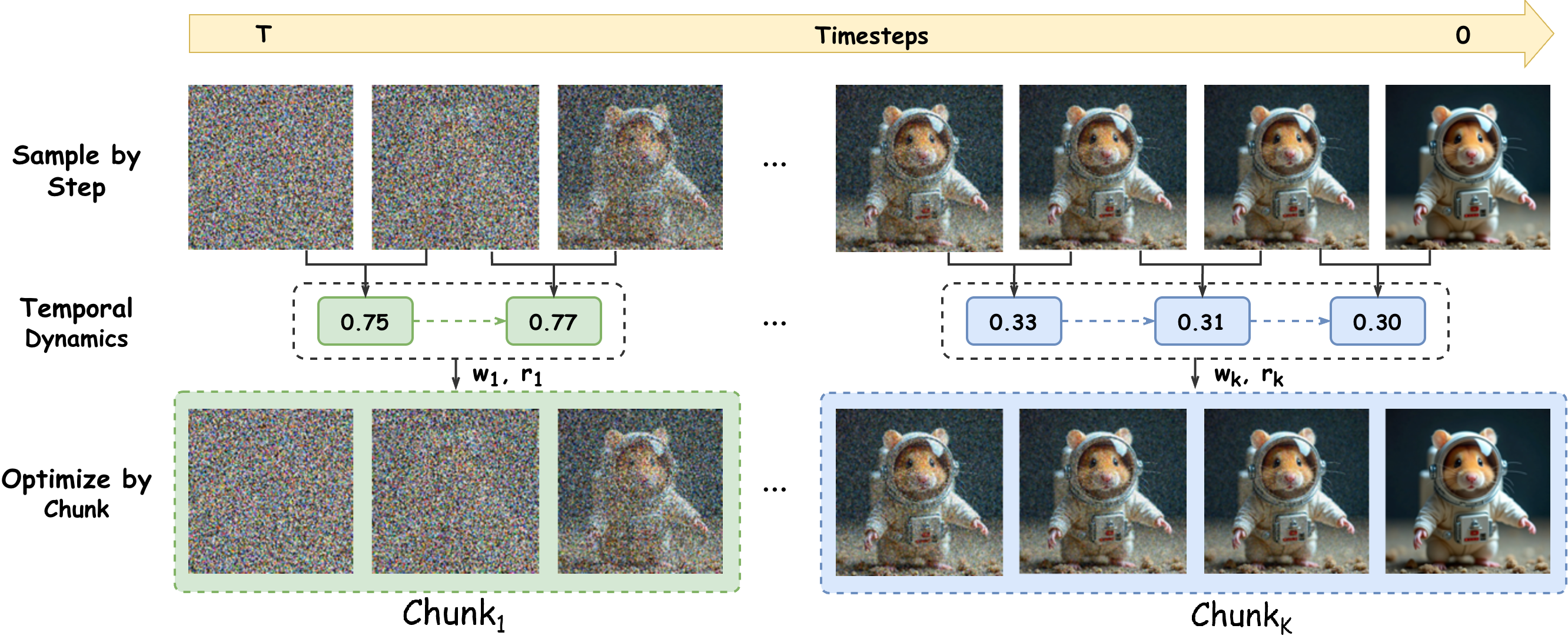}
\caption{Overall framework of GCPO. GCPO shifts the policy optimization from the step level to the chunk level based on temporal-dynamic-guided chunking, using a principled chunk-level importance ratio $r$. It also introduces an optional weighted sampling strategy that assigns a sampling weight $w$ to each chunk.}
\label{figure_framework}
\vskip -0.1in
\end{figure*}

Building on these insights, we introduce Group Chunking Policy Optimization (GCPO), the first chunk-level RL approach in post-training flow matching for T2I generation. As illustrated in \Cref{figure_framework}, our key innovation lies in shifting the policy optimization paradigm from the step level to the chunk level and treating each chunk as a single action with a principled chunk-level importance ratio. In addition, we incorporate an optional weighted sampling strategy to further enhance GCPO's performance.

Our contributions can be summarized as follows:

\begin{itemize}
    \item We pioneer chunk-level RL for T2I generation. By treating consecutive flow matching steps as a single action, we mitigate the adverse effects of inaccurate advantage attribution presented in GRPO methods, leading to better training stability.  
    
    \item We propose GCPO, a novel chunk-level policy optimization approach in post-training flow matching for T2I generation, which integrates chunk-level importance ratio with temporal-dynamic-guided chunking. An optional weighted sampling strategy is introduced to further boost performance.
    
    \item Extensive experiments demonstrate that GCPO achieves superior performance on both preference alignment and standard T2I benchmarks, significantly surpassing prior RL approaches.
\end{itemize}

\section{Related Work}

\subsection{Action Chunking}
Action chunking \cite{zhao2023learning, lai2025action} has been widely applied to robotics \cite{chi2025diffusion}. This approach mitigates compounding error and non-Markovian noise in human demonstrations by jointly predicting a sequence of future actions as a single action, which enables smoother and more stable rollouts. Recently, it has also proven effective in vision-language-action models \cite{black2024pi_0, black2025pi_} and in RL \cite{li2025reinforcement}. These successes suggest that action chunking accelerates value propagation and stabilizes long-horizon prediction.

\subsection{Reinforcement Learning for Flow-matching-based Image Generation} 

Diffusion and flow matching models \citep{ho2020denoising, rombach2022high, podellsdxl, lipmanflow, batifol2025flux, wu2025qwen, wang2026elastic} have become one of the dominant paradigms for T2I generation. Early works \citep{xu2023imagereward, blacktraining, fan2023dpok} introduced RL into diffusion models through policy gradient optimization. Preference-based methods \citep{wallace2024diffusion, sundiffusion, sun2025generalizing} were later developed, achieving competitive alignment without explicit reward modeling.

More recently, GRPO \citep{shao2024deepseekmath,sun2025reinforcement} has attracted attention as an efficient alternative. Dance-GRPO \citep{xue2025dancegrpo} and Flow-GRPO \citep{liu2026flow} pioneered the use of GRPO for T2I generation, unifying diffusion and flow matching through an SDE-based reformulation. MixGRPO \citep{li2025mixgrpo} further improved efficiency via a mixed ODE–SDE paradigm. TempFlow-GRPO \citep{he2025tempflow} introduced temporal-aware weighting across denoising steps. Pref-GRPO \citep{wang2025pref} identified the issue of illusory advantage and reformulated the optimization objective as pairwise preference fitting. BranchGRPO \citep{li2025branchgrpo} restructured the rollout process into a branching tree, amortizing computation across shared prefixes.

In contrast to these works, our approach aims to mitigate the adverse effects of one key issue in GRPO methods: inaccurate advantage attribution. By introducing chunk-level policy optimization guided by the inherent temporal dynamics of flow matching, we enhance GRPO from the perspective of policy optimization granularity and stabilize its training. Note that the key difference of our approach from some process-reward methods, such as \citep{deng2026densegrpo}, lies in that we do not introduce any process rewards. Instead, we still assign the same outcome reward but shift the optimization from the step level to chunk level. Additional empirical results are provided in \Cref{additional_baselines_app}.

\section{Preliminary}

\subsection{Flow Matching}

Suppose that $x_0\sim\mathbb{X}_0$ is a data sample from the true distribution, and $x_1\sim\mathbb{X}_1$ is a noise sample. Following \cite{liuflow}, the intermediate noisy samples $x_t$ can be expressed as:
\begin{equation}
\label{preliminaryeq1}
    x_t = (1-t)x_0 + tx_1,
\end{equation}
where $t\in[0,1]$ denotes the noise level. Then, flow matching aims to directly regress the estimated velocity field $\hat{v}_\theta(x_t,t)$ by minimizing the objective function \cite{lipman2022flow}:
\begin{equation}
\label{preliminaryeq2}
    \mathcal{L}_{\text{FM}}(\theta) = \mathbb{E}_{t, x_0\sim\mathbb{X}_0, x_1\sim\mathbb{X}_1}[\|v-\hat{v}_{\theta}(x_t,t)\|_2^2],
\end{equation}
where $v=x_1-x_0$ represents the target velocity field. Furthermore, a deterministic Ordinary Differential Equation (ODE) is utilized to model the forward process of flow matching:
\begin{equation}
\label{preliminaryeq3}
    dx_t=\hat{v}_\theta(x_t,t)dt.
\end{equation}

\subsection{GRPO on Flow Matching}

As an RL algorithm, GRPO \cite{guo2025deepseek, shao2024deepseekmath} effectively eliminates the need for an additional critic model by estimating the baseline through group-wise relative rewards. In line with the settings of DDPO \cite{blacktraining}, GRPO is also applied in flow matching. Given a group of $G$ images $\{x_0^i\}_{i=1}^G$ generated from the same prompt $c$, the advantage corresponding to the $i$-th sample is formulated as: 
\begin{equation}
    {A}_t^i=\frac{r(x_0^i, c)-\text{mean}(\{r(x_0^j, c)\}_{j=1}^G)}{\text{std}(\{r(x_0^j, c)\}_{j=1}^G)}.
\label{preliminaryeq4}
\end{equation}
Notice that ${A}_t^i$ always keeps the same value for any step $t$. For simplicity, we neglect the subscript and denote it as ${A}^i$. The policy is updated by maximizing the following GRPO objective:
\begin{equation}
\begin{aligned}
    J(\theta) &= \mathbb{E}_{c, \{x^i\}_{i=1}^G} \Bigg[\frac{1}{G} \frac{1}{T} \sum_{i=1}^{G} \sum_{t=1}^{T} \\
    &\big(\min \left(r_t^i (\theta)A^i, \text{clip}\left(r_t^i (\theta), 1- \epsilon, 1 + \epsilon \right)A^i\right) \\
    &- \beta D_{KL}\left(\pi_\theta || \pi_{ref}\right) \big)\Bigg],
\end{aligned}
\label{preliminaryeq5}
\end{equation}
where $r_t^i$ denotes the step-level importance ratio:
\begin{equation}
\begin{aligned}
        r^i_t(\theta) = \frac{p_\theta (x^i_{t-1} | x^i_t, c)}{p_\text{old} (x^i_{t-1} | x^i_t, c)}.
\end{aligned}
\label{preliminaryeq6}
\end{equation}
Furthermore, to meet the exploration requirement of RL, Flow-GRPO \cite{liu2026flow} and Dance-GRPO \cite{xue2025dancegrpo} introduce stochasticity into flow matching by transforming the deterministic ODE into an equivalent Stochastic Differential Equation (SDE):
\begin{equation}
    dx_t = \big(v_{\theta}(x_t,t)+\frac{\sigma_t^2}{2t}(x_t+(1-t)v_{\theta}(x_t,t))\big)dt + \sigma_t d w_t,
\label{preliminaryeq7}
\end{equation}
where $dw_t$ represents the increments of the Wiener process and $\sigma_t$ controls the stochasticity.

\section{Method}

In this section, we begin by reformulating the RL objective from the step level to chunk level and demonstrating why this approach improves GRPO in \Cref{From Step-level to Chunk-level Optimization}. Next, we explain how to guide chunking process using the distinct temporal dynamics of flow matching in \Cref{Chunking Guided by Inherent Temporal Dynamics}. Finally, we present our proposed GCPO along with an optional weighted sampling strategy in \Cref{Group Chunking Policy Optimization}.

\subsection{From Step-level to Chunk-level Policy Optimization}\label{From Step-level to Chunk-level Optimization}

We begin by revisiting the two-trajectory example in \Cref{figure_motivation}. With the step-level GRPO objective defined in \Cref{preliminaryeq5}, the optimization objective for this example is (omitting the KL term for simplicity):
\begin{equation}
\begin{aligned}
    J(\theta) = &\frac{1}{2} \frac{1}{2} \sum_{i=1}^2 \sum_{t=1}^2 \min  \\
    &\left(r_t^i \left(\theta \right)A^i, \text{clip}\left(r_t^i \left(\theta \right), 1- \epsilon, 1 + \epsilon \right)A^i\right).
\end{aligned}
\label{methodeq1}
\end{equation}
As previously discussed, GRPO's uniform advantage assignment introduces inaccurate advantage attribution, resulting in training instability. To address this issue, the core principle of chunk-level policy optimization is to group consecutive steps into a chunk and treat it as a single, atomic action. In the example case, where the steps $t=1$ and $t=2$ form a chunk, the optimization objective becomes to:
\begin{equation}
\begin{aligned}
    J(\theta) =& \frac{1}{2} \sum_{i=1}^2 \min \\
    &\left(r^i \left(\theta \right)A^i, \text{clip}\left(r^i \left(\theta \right), 1- \epsilon, 1 + \epsilon \right)A^i\right),
\end{aligned}
\label{methodeq2}
\end{equation}
where the importance ratio $r$ is redefined over the chunk-level likelihood:
\begin{equation}
    r^i(\theta) = \left( \prod_{t=1}^2 \frac{p_\theta \left( x_{t-1}^i | x_t^i, c \right)}{p_{\text{old}} \left( x_{t-1}^i | x_t^i, c \right)} \right) ^{\frac{1}{2}}.
\label{methodeq3}
\end{equation}
Comparing \Cref{methodeq2} and \Cref{methodeq3} with \Cref{methodeq1} and \Cref{preliminaryeq6}, we observe that the policy optimization paradigm shifts to the chunk level via a principled chunk-level importance ratio. This smooths the misleading fluctuations caused by the inaccurate advantage attribution. A detailed theoretical analysis is provided in \Cref{Theoretical Analysis}.

Building on this insight, we formally formulate chunk-level policy optimization as follows. Given an image generation trajectory:
\begin{equation}
    (x_{T}, x_{T-1}, \cdots, x_2, x_1, x_0)^i,
\label{methodeq4}
\end{equation}
we partition it into $K$ distinct chunks: \footnote{we neglect $x_0^i$ as it has no preceding transition to $x_{-1}^i$.}
\begin{equation}
\begin{aligned}
    &\{ch_1, \cdots, ch_{K}\}^i \\
    =&\{(x_{T}, \cdots, x_{T-cs_1+1}),\cdots, (x_{cs_K}, \cdots, x_1)\}^i,
\end{aligned}
\label{methodeq5}
\end{equation}
where $cs_j$ denotes the chunk size (the number of consecutive steps within a chunk) of the $j$-th chunk $ch_j$, and:
\begin{equation}
    \sum_{j=1}^{k} cs^i_j = T.
\end{equation}

\begin{figure}[ht]
  \centering
    \includegraphics[width=1.0\linewidth]{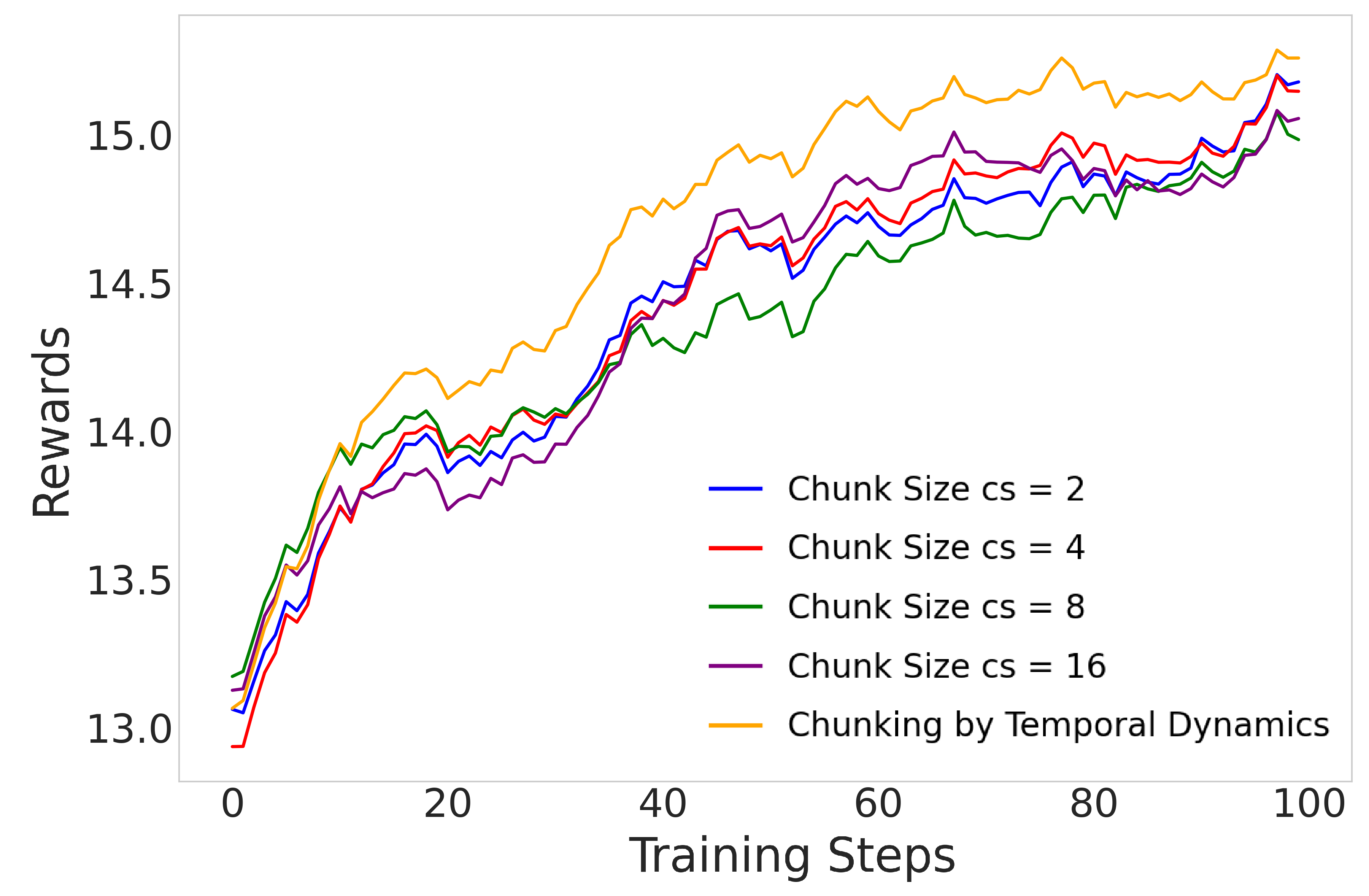}
     \vskip -0.06in
    \caption{
      Performance varies with different chunkings. The chunking with temporal dynamics is our method.
    }
    \label{figure_toy_experiment} 
    \vskip -0.1in
\end{figure} 

The chunk-level optimization objective is then:
\begin{equation}
\begin{aligned}
    J(\theta) &= E_{c, \{x^i\}_{i=1}^G} \bigg[\frac{1}{G} \frac{1}{K} \sum_{i=1}^{G} \sum_{j=1}^{K} \\
    &\big(\min \left(r_j^i \left(\theta \right)A^i, \text{clip}\left(r_j^i \left(\theta \right), 1- \epsilon, 1 + \epsilon \right)A^i\right) \\
    &- \beta D_{KL}\left(\pi_\theta || \pi_{ref}\right) \big)\bigg],
\end{aligned}
\label{methodeq6}
\end{equation}
where we redefine the importance ratio $r_j^i(\theta)$ based on the chunk-level likelihood:
\begin{equation}
    r_j^i(\theta) = \left( \prod_{t \in ch_j} \frac{p_\theta \left( x_{t-1}^i | x_t^i, c \right)}{p_{\text{old}} \left( x_{t-1}^i | x_t^i, c \right)} \right) ^{\frac{1}{cs_j}}.
\label{methodeq7}
\end{equation}
Note that we normalize the importance ratio by the chunk size to ensure stability, following \cite{zheng2025group}.

It is evident that various chunkings exist. For example, setting $K=1$ treats the entire generation trajectory as a single chunk, and the policy optimization further shifts from chunk-level to sequence-level, similar to \cite{zheng2025group}. Conversely, setting $K=T$ forces $cs_j=1$, reverting the policy optimization back to step-level GRPO. Consequently, in the next section, we will discuss the principles that guide chunking.

\subsection{Chunking Guided by Inherent Temporal Dynamics}\label{Chunking Guided by Inherent Temporal Dynamics}

Before diving into the detailed analysis, we conducted a preliminary experiment where all chunks shared an equal size $cs_1 = cs_2 \cdots = cs_k$. As shown in \Cref{figure_toy_experiment}, performance varies with chunkings, indicating that the choice of chunking is non-trivial.

\begin{figure*}
\centering
\includegraphics[width=1.0\textwidth]{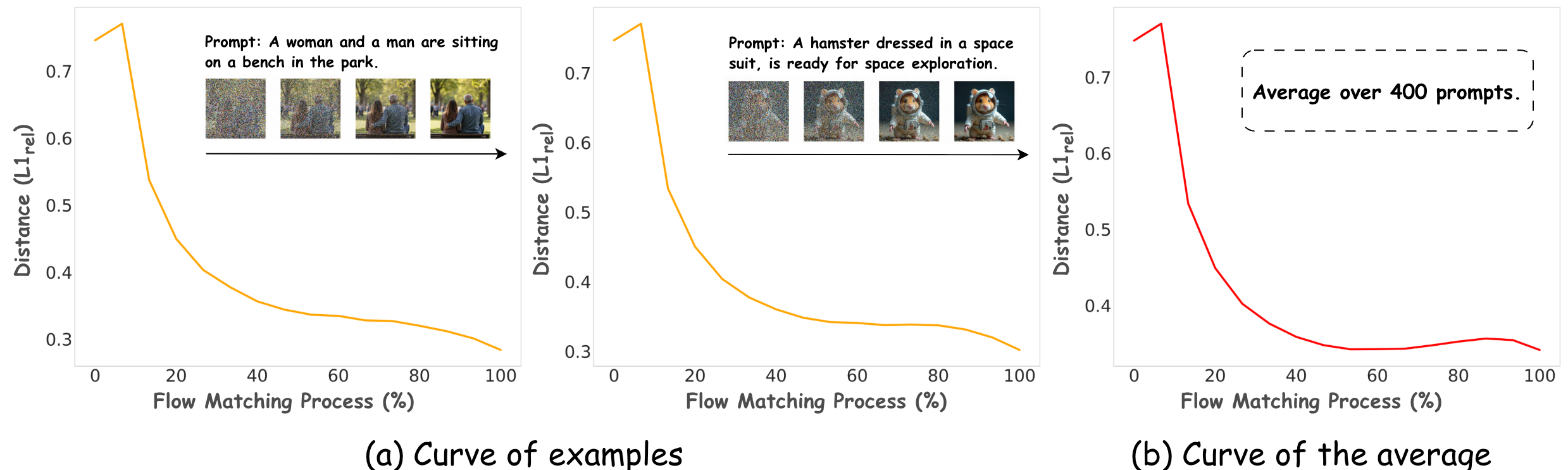}
\caption{Illustration of the prompt-invariant yet step-dependent temporal dynamics of flow matching, where y-axis denotes the relative $L1$ distance $L1_\text{rel} (x, t)$ in \Cref{methodeq8}.}
\label{figure_relative_distance}
\vskip -0.1in
\end{figure*}

We attribute this variation to the inherent temporal dynamics of flow matching. Unlike autoregressive LLMs, flow matching involves time-dependent dynamics during generation, where different steps contribute unequally to the final image. To better understand this, following \cite{wimbauer2024cache, liu2025timestep}, we illustrate the relative $L1$ distance $L1_\text{rel} (x, t)$ throughout the generation process:
\begin{equation}
    L1_\text{rel} (x, t) = \frac{\lVert x_t - x_{t-1} \rVert_1}{\lVert x_t \rVert_1}.
\label{methodeq8}
\end{equation}
The relative $L1$ distance quantifies the rate of change in the latent space during generation. As shown in \Cref{figure_relative_distance}, $L1_\text{rel} (x, t)$ exhibits prompt-invariant yet step-dependent patterns. Crucially, it naturally segments the generation trajectory into meaningful chunks, where a large $L1_\text{rel} (x, t)$ indicates rapid latent changes, while a small value implies that adjacent latents are similar. Based on this observation, we argue that steps with similar temporal dynamics (e.g., $L1_\text{rel} (x, t)$) should be grouped into the same chunk, whereas those with distinct dynamics should be separated into different chunks.

Consequently, we guide the chunking process using temporal dynamics (e.g., $L1_\text{rel} (x, t)$), aligning the chunk-level policy optimization with the intrinsic temporal dynamics of flow matching. In the next section, we will discuss the chunking implementation details.

\subsection{Group Chunking Policy Optimization}\label{Group Chunking Policy Optimization}

We now present Group Chunking Policy Optimization (GCPO), which integrates chunk-level policy optimization with temporal-dynamic-guided chunking.

Specifically, for a given image generation trajectory in \Cref{methodeq4}, we first capture temporal dynamics (e.g., $L1_\text{rel} (x, t)$ following \Cref{methodeq8}), and then partition the trajectory into chunks according to them. Treating each chunk as a single action, the policy optimization follows the chunk-level objective in \Cref{methodeq6}. The whole framework is shown in \Cref{figure_framework}. 

In practice, we employ an adaptive chunking strategy by iteratively analyzing the derivative of the relative $L1$ distance for each generation trajectory. Specifically, we first compute the first-order derivatives for each step and group consecutive steps sharing the same sign into chunks. In cases where the derivative signs are uniform across the entire trajectory, we split the sequence at its midpoint. We then recursively apply this process to each resulting chunk using progressively higher-order derivatives (second-order, third-order, etc.). The recursion terminates for a given chunk when its size becomes sufficiently small, and the process concludes once all chunks meet this termination condition. 

It is important to note that, unlike other methods that attempt to resolve inaccurate advantage attribution by introducing process rewards, our approach retains the equal assignment of final rewards across all steps (chunks). However, we aim to mitigate the negative influence of this issue by optimizing the policy at the chunk level with the principled chunk-level importance ratio. This effectively smooths the unexpected misleading gradient fluctuations caused by inaccurate advantage attribution.

Furthermore, we propose an optional weighted sampling strategy to further enhance GCPO. Following Dance-GRPO \cite{xue2025dancegrpo}, in practice, we select only a subset of chunks (e.g., with fraction $0.5$) from each trajectory for training. However, instead of uniform sampling, we assign a sampling weight $w$ to each chunk:
\begin{equation}
    w(ch_j) = \frac{\frac{1}{cs_j} \sum_{t \in ch_j} L1_\text{rel} \left(x, t\right )}{\sum_{j=1}^K\frac{1}{cs_j} \sum_{t \in ch_j} L1_\text{rel} \left(x, t\right )}.
\label{methodeq9}
\end{equation}
The motivation stems from our ablation studies in \Cref{Training on Specific Chunks}, where we observed varying training contributions from different chunks. As shown in \Cref{figure_relative_distance}, this strategy biases the sampling toward high-noise regions. However, although this strategy enhances GCPO in terms of preference alignment, its overall impact on image quality is mixed and nuanced, as discussed in \Cref{Weighted Sampling Strategy}.

\begin{table*}[t]
\setlength{\tabcolsep}{3mm} 
\centering
\small
\begin{threeparttable}[b]
\caption{Performance on GenEval and DPG.}
\begin{tabular}{lcccccccc}
\toprule
\multirow{2}{*}{\textbf{Methods}} & \multicolumn{7}{c}{\textbf{GenEval}} & \multicolumn{1}{c}{\textbf{DPG}} \\
\cmidrule(lr){2-9} 

& Single Obj. & Two Obj. & Counting & Colors & Position & Color Attr. & Overall & Overall \\
\midrule
Flux & 0.99 & 0.83 & 0.71 & 0.75 & 0.24 & 0.44 & 0.66 & 84.00 \\
\midrule
Dance-GRPO & \cellcolor{C1!50!white} 1.00 & \cellcolor{C1!50!white}0.86 & 0.71 & 0.78 & 0.22 & 0.46 & 0.67 & 85.17\\
Flow-GRPO & 0.99 & 0.84 & 0.71 & 0.79 & 0.21 & 0.46 & 0.67 & 85.05\\
\midrule
GCPO w/o ws & 0.99 & 0.85 & \cellcolor{C1!50!white}0.75 & \cellcolor{C1!50!white}0.81 & 0.21 & \cellcolor{C1!50!white}0.51 & \cellcolor{C1!50!white}0.69 & \cellcolor{C1!50!white} 86.60\\
GCPO w/ ws & 0.98 & 0.82 & 0.73 & 0.76 &\cellcolor{C1!50!white}0.27 & 0.48 & 0.67 & 85.14\\
\bottomrule
\end{tabular}
\label{table_geneval}
\begin{tablenotes}
  \item[1] Here and in following tables, the `ws' refers to the weighted sampling strategy.
\end{tablenotes}
\end{threeparttable}
\vskip -0.1in
\end{table*}

\section{Experiments}

\subsection{Experiment Setup}

\textbf{Training Settings. } We adopt Dance-GRPO \cite{xue2025dancegrpo} and Flow-GRPO \cite{liu2026flow} as baselines, conducting experiments with FLUX.1 Dev \cite{flux2024} as our base model. Comparison of additional baselines is provided in \Cref{additional_baselines_app}. HPDv2.1 \cite{wu2023human} serves as the dataset, while HPSv3 \cite{ma2025hpsv3} and CLIP \cite{radford2021learning} are used as the primary reward models for the preference alignment task and the standard T2I benchmarks, respectively. In our ablation studies \Cref{Reward Models}, we additionally validate our approach using Pick Score \cite{kirstain2023pick} as the reward model. Further training details are provided in \Cref{Details for Training}.  

\textbf{Evaluation Details. } We evaluate performance on both preference alignment and standard T2I benchmarks. For preference alignment, we use HPSv3 \cite{ma2025hpsv3} and ImageReward \cite{xu2023imagereward} as in-domain and out-of-domain evaluation metrics, respectively, on the HPDv2.1 \cite{wu2023human} test set. For the standard T2I benchmark, we report results on GenEval \cite{ghosh2023geneval} and DPG \cite{hu2024ella}. All evaluations utilize the hybrid inference strategy from \cite{li2025mixgrpo}, which has proven effective in mitigating reward hacking. Additionally, we conduct a user study to assess human preferences. More evaluation details are provided in \Cref{Details for Evaluation}.   

\subsection{Main Results}

\Cref{table_preference_alignment} presents results for preference alignment, and \Cref{table_geneval} details performance on the GenEval and DPG benchmarks. GCPO consistently outperforms both the base model and the baselines. In preference alignment, our approach achieves significant relative gains of up to $43\%$ over the baseline across both in-domain and out-of-domain metrics. On GenEval and DPG, our approach achieves the strongest performance, with relative gains up to three times larger than those of the baselines, despite GenEval being a notoriously challenging benchmark when relying solely on CLIP rewards. 
Notably, while the weighted sampling strategy enhances preference alignment, it yields mixed effects on standard T2I benchmarks, a phenomenon we further analyze in \Cref{Weighted Sampling Strategy}.

Qualitative comparisons in \Cref{figure_quality_main} and \Cref{Main Visualization} further highlight GCPO's improvements in image quality. Our method generates outputs that align more closely with human aesthetic preferences, exhibiting stronger lighting contrast, more vivid colors, and finer details. Notably, the contrast between columns $3$ and $4$  underscores the importance of chunking in relation to the inherent temporal dynamics of flow matching, which substantially elevates the aesthetic style and structure of the images.

We also conducted a user study to assess human preferences. Reviewers were tasked with selecting the best image from 3 alternatives generated by Dance-GRPO, GCPO without weighted sampling strategy, and GCPO with weighted sampling strategy, across $40$ prompts. The results in \Cref{table_user_study} show that the GCPO variants were preferred by human reviewers $72.5\%$ of the time, which further demonstrates the significant superiority of our approach in aligning with human preferences. More details are provided in \Cref{Details for User Study}.

\begin{table}[t]
  \setlength{\tabcolsep}{3mm} 
  \caption{Performance on Preference Alignment.}
  \vskip -0.07in
  \label{table_preference_alignment}
  \begin{center}
    \begin{small}
      \begin{threeparttable}[b]
        \begin{tabular}{lcc}
          \toprule
          \textbf{Methods} & \textbf{HPSv3} & \textbf{ImageReward} \\
          \midrule
          Flux & 13.804 & 1.086 \\
          \midrule
          Dance-GRPO & 15.080 & 1.141 \\
          Flow-GRPO & 14.900 & 1.135 \\
          \midrule
          GCPO w/o ws  & 15.236 & 1.147 \\
          GCPO w/ ws  &\cellcolor{C1!50!white} 15.373 &\cellcolor{C1!50!white} 1.149 \\
          \bottomrule
        \end{tabular}
        \end{threeparttable}
    \end{small}
  \end{center}
%   \vskip -0.19in
\end{table}

\begin{table}[t]
  \setlength{\tabcolsep}{5mm} 
  \caption{Results of User Study.}
   \vskip -0.07in
  \label{table_user_study}
  \begin{center}
    \begin{small}
      \begin{threeparttable}[b]
        \begin{tabular}{lc}
          \toprule
          \textbf{Methods} & \textbf{Win Rate} \\
          \midrule
          Dance-GRPO & 0.275 \\
          \midrule
          GCPO w/o ws & 0.350 \\
          GCPO w/ ws & \cellcolor{C1!50!white} 0.375 \\
          \bottomrule
        \end{tabular}
        \end{threeparttable}
    \end{small}
  \end{center}
  \vskip -0.1in
\end{table}

\begin{figure*}
\centering
\includegraphics[width=0.93\textwidth]{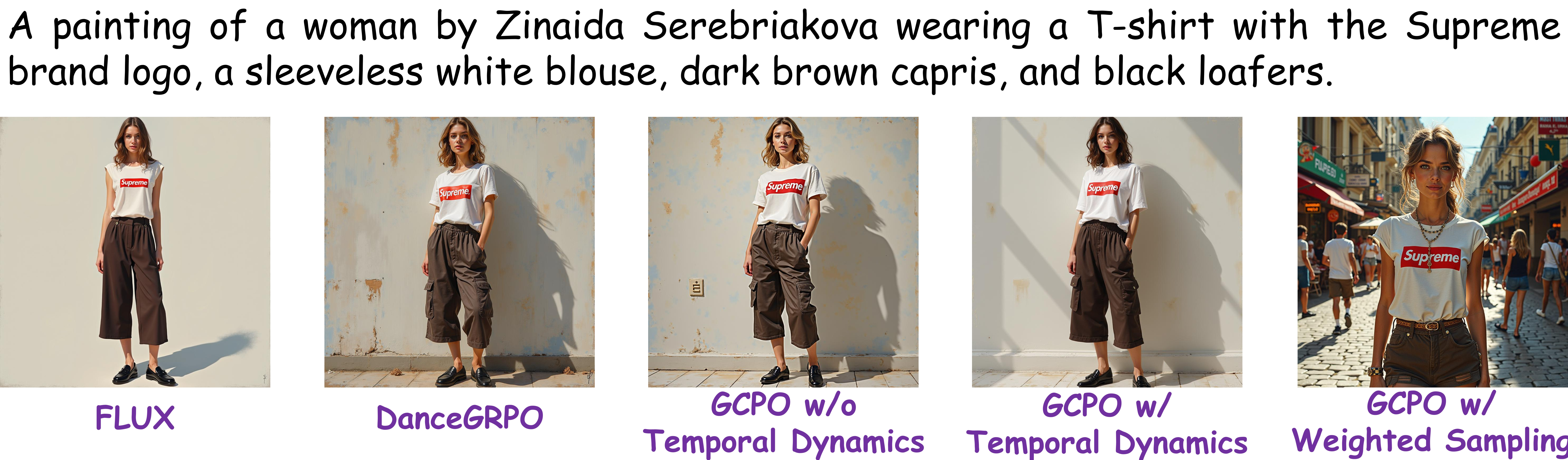}
\caption{A failure case of the weighted sampling strategy. The strategy erroneously alters the image structure within high-noise regions, resulting in the poorest variant.}
\label{figure_failure_main}
\vskip -0.05in
\end{figure*}

\subsection{Ablation Study}

Additional ablation studies are provided in \Cref{additional_ablation_study_app}. 

\begin{figure}[t]
%   \vskip 0.05 in
  \centering
    \includegraphics[width=0.92\linewidth]{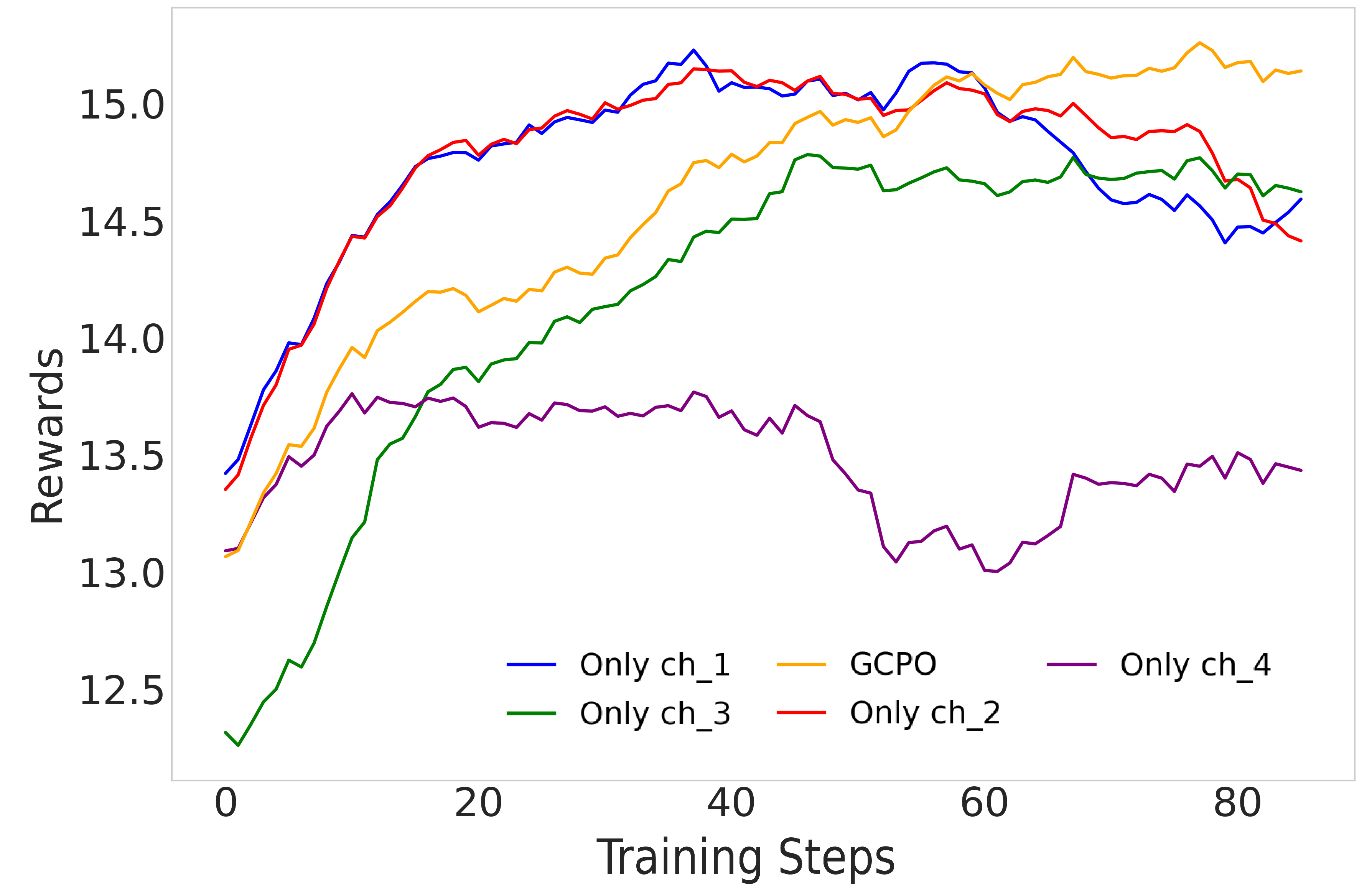}
    % \vskip -0.06in
    \caption{
      Results of training on specific chunks. Lower chunk indices correspond to high-noise regions.
    }
    % \vskip -0.1in
    \label{figure_specific_chunk} 
\vskip -0.2in
\end{figure}

\textbf {Training on Specific Chunks. } \label{Training on Specific Chunks} We analyze the training contributions of distinct temporal regions by training GCPO on individual chunks only. For simplicity, we fix the chunk size and divide the generation trajectory to $K=4$ chunks. Results in \Cref{figure_specific_chunk} indicate that while high-noise chunks yield larger improvements than low-noise chunks, they suffer from training instability (e.g., after $60$ steps). This observation motivated the weighted sampling strategy in \Cref{methodeq9}, which adaptively emphasizes high-noise chunks to accelerate training while retaining low-noise chunks to ensure stability.      

\textbf {Chunk Setting. } In general, there are two ways of constructing chunks. One is the adaptive chunking strategy described in \Cref{Group Chunking Policy Optimization}, which we adopt as our default implementation. The other is keeping the chunking fixed during the training, despite different prompts and temporal dynamics. Results in \Cref{chunk setting ablation} show that chunk-level optimization consistently outperforms standard step-level GRPO. Moreover, temporal-dynamics-guided chunking outperforms fixed chunk size, underscoring the importance of aligning the optimization process with the intrinsic temporal structure of flow matching. Note that the weighted sampling strategy is disabled here for fairness.        

\textbf{Weighted Sampling Strategy. } \label{Weighted Sampling Strategy} 
As shown in Tables \ref{table_preference_alignment} and \ref{table_geneval}, the optional weighted sampling strategy improves preference alignment but slightly reduces standard T2I benchmark performance. Careful qualitative analysis reveals a trade-off: while this strategy accelerates policy optimization, it can destabilize image structures in high-noise regions, occasionally leading to semantic collapse. A failure example is depicted in \Cref{figure_failure_main}. Although all methods struggle with this challenging prompt (e.g., Dance-GRPO fails to generate the attribute `sleeveless'), the weighted sampling strategy further degrades the overall image structure, producing the worst outcome by omitting the item `black loafers' entirely and only partially rendering the `capris'. This demonstrates the mixed and nuanced effects of the strategy. More qualitative comparison is provided in \Cref{Visualization Comparison of the Weighted Sampling Strategy}.

\begin{table}[t]
\vskip -0.1in
  \setlength{\tabcolsep}{3mm} 
  \caption{Ablation Results of Different Chunk Settings.}
  \vskip -0.05in
  \label{chunk setting ablation}
  \begin{center}
    \begin{small}
      \begin{threeparttable}[b]
        \begin{tabular}{lcc}
          \toprule
          \textbf{Methods} & \textbf{Chunking} & \textbf{HPSv3} \\
          \midrule
          Flux & - & 13.804 \\
          \midrule
          Dance-GRPO & - & 15.080 \\
          GCPO (default) & adaptive & \cellcolor{C1!50!white} 15.236 \\
          \midrule
          \multirow{4}{*}{GCPO (fixed)} & [2, $\cdots$, 2] & 15.115 \\
           & [4, 4, 4, 4] & 15.078 \\
           & [8, 8] & 15.173 \\
           & [16] & 15.142 \\
          \bottomrule
        \end{tabular}
        \end{threeparttable}
    \end{small}
  \end{center}
  \vskip -0.1in
\end{table}

\begin{table}[t]
    % \vskip -0.1in
  \caption{Ablation Results on Different Reward Models.}
  \vskip -0.05in
  \label{table_pickscore}
  \begin{center}
    \begin{small}
      \begin{threeparttable}[b]
        \begin{tabular}{lccc}
          \toprule
          \textbf{Model} & \textbf{PickScore} & \textbf{HPSv3} & \textbf{ImageReward} \\
          \midrule
          Flux & 22.643 & 13.804 & 1.086 \\
          \midrule
          Dance-GRPO & 23.427 & 14.612 & 1.208 \\
          Flow-GRPO & 23.335 & 14.610 & 1.186\\
          \midrule
          GCPO w/o ws & 23.442 & 14.810 & 1.222 \\
          GCPO w/ ws & \cellcolor{C1!50!white} 23.476 & \cellcolor{C1!50!white} 14.913 & \cellcolor{C1!50!white} 1.233 \\
          \bottomrule
        \end{tabular}
        \end{threeparttable}
    \end{small}
  \end{center}
   \vskip -0.2in
\end{table}

\textbf{Reward Models. } \label{Reward Models} Finally, we test GCPO's robustness across different reward models. We replace HPSv3 with Pick Score \citep{shukor2025smolvla} as our reward model. Results in \Cref{table_pickscore} confirm that our method consistently outperforms step-level GRPO regardless of the reward model used, demonstrating its generalization capabilities.

\section{Conclusion}

In this paper, we propose GCPO, the first chunk-level policy optimization approach for post-training flow matching in T2I generation. Guided by the temporal dynamics of flow matching, GCPO aggregates consecutive steps into a chunk and optimizes shifts the policy optimization to the chunk level, achieving consistent improvements over GRPO.

Despite its strong performance, several limitations remain. For example, exploring how to combine heterogeneous rewards across different chunks (e.g., employing different reward models for high- vs. low-noise regions) could unlock further improvements.

\section*{Acknowledgements}

This work was partially supported by the China Postdoctoral Science Foundation under Grant No. 2025M781490, the Natural Science Foundation of Shenzhen under Grant Nos. JCYJ20230807111604008 and JCYJ20240813112007010, the Natural Science Foundation of Guangdong Province under Grant No. 2024A1515010003 and National Key Research and Development Program of China under Grant No. 2022YFB4701400.

\section*{Impact Statement}

This paper presents GCPO, a chunk-level reinforcement learning approach designed to enhance the visual quality and preference alignment of text-to-image flow matching models. Our work aims to advance the capabilities of generative AI as a tool for creative expression and visual synthesis. However, we acknowledge that the significant improvements in structural coherence, lighting, and photorealism demonstrated by GCPO could potentially be misused to generate misleading content or deepfakes that are difficult to distinguish from reality. Furthermore, since our optimization relies on preference reward models, there is an inherent risk of amplifying biases present in these evaluators or the underlying training data. We encourage the community to deploy such advanced post-training methods alongside robust safety filters, watermarking technologies, and diverse reward modeling to mitigate these risks.

% % In the unusual situation where you want a paper to appear in the
% % references without citing it in the main text, use \nocite
% \nocite{langley00}

\bibliography{icml2026}
\bibliographystyle{icml2026}

%%%%%%%%%%%%%%%%%%%%%%%%%%%%%%%%%%%%%%%%%%%%%%%%%%%%%%%%%%%%%%%%%%%%%%%%%%%%%%%
%%%%%%%%%%%%%%%%%%%%%%%%%%%%%%%%%%%%%%%%%%%%%%%%%%%%%%%%%%%%%%%%%%%%%%%%%%%%%%%
% APPENDIX
%%%%%%%%%%%%%%%%%%%%%%%%%%%%%%%%%%%%%%%%%%%%%%%%%%%%%%%%%%%%%%%%%%%%%%%%%%%%%%%
%%%%%%%%%%%%%%%%%%%%%%%%%%%%%%%%%%%%%%%%%%%%%%%%%%%%%%%%%%%%%%%%%%%%%%%%%%%%%%%
\newpage
\appendix
\onecolumn

\section{Theoretical Analysis}\label{Theoretical Analysis}

Here, we present the superiority of chunk-level policy optimization from the perspective of gradient weights, which demonstrates why it offers a more stable training.

To begin with, the gradient of the GRPO's objective in \Cref{preliminaryeq5} can be derived as follows (clipping and KL are omitted for brevity)\footnote{The omission of clipping and KL terms makes no difference to the gradient derivation. When the clipping occurs, the gradient of the clipped term is. Regarding KL, it is identical for both GCPO and GRPO, therefore its gradient contribution is the same.}:
\begin{equation}
\begin{aligned}
    \nabla_\theta J_\text{GRPO}(\theta) &= \nabla_\theta E_{c, x} \left[\frac{1}{G} \frac{1}{T} \sum_{i=1}^{G} \sum_{t=1}^{T} r_t^i \left(\theta \right)A^i \right] \\\\\\
    &= \nabla_\theta E_{c, x} \left[\frac{1}{G} \sum_{i=1}^{G} A^i \frac{1}{T} \sum_{t=1}^{T} r_t^i \left(\theta \right)
    \cdot \nabla_\theta \text{log} \left( r_t^i \left(\theta \right) \right) \right].
\end{aligned}
\end{equation}
Given \Cref{preliminaryeq6}, this simplifies to:
\begin{equation}
\begin{aligned}
    \nabla_\theta J_\text{GRPO}(\theta) &= \nabla_\theta E_{c, x} \left[\frac{1}{G} \sum_{i=1}^{G} A^i \frac{1}{T} \sum_{t=1}^{T} \frac{p_\theta (x^i_{t-1} | x^i_t, c)}{p_\text{old} (x^i_{t-1} | x^i_t, c)}
    \cdot \nabla_\theta \text{log} \left( p_\theta (x^i_{t-1} | x^i_t, c) \right) \right].
\end{aligned}
\end{equation}
In comparison, the gradient of the chunk-level policy optimization objective in \Cref{methodeq6} is derived below. For convenience, we denote the chunk-level importance ratio from \Cref{methodeq7} as:
\begin{equation}
s_j^i(\theta) = \left( \prod_{t \in ch_j} \frac{p_\theta \left( x_{t-1}^i | x_t^i, c \right)}{p_{\theta_\text{old}} \left( x_{t-1}^i | x_t^i, c \right)} \right) ^{\frac{1}{cs_j}},
\end{equation}
Then the gradient becomes:
\begin{equation}
\begin{aligned}
    \nabla_\theta J_\text{GCPO}(\theta) &= \nabla_\theta E_{c, x} \left[\frac{1}{G} \frac{1}{K} \sum_{i=1}^{G} \sum_{j=1}^{K} s_j
    ^i \left(\theta \right)A^i \right] \\
    &= \nabla_\theta E_{c, x} \left[\frac{1}{G} \sum_{i=1}^{G} A^i \frac{1}{K} \sum_{j=1}^{K} s_j^i \left(\theta \right)
    \cdot \nabla_\theta \text{log} \left( s_j^i \left(\theta \right) \right) \right] \\
    &= \nabla_\theta E_{c, x} \frac{1}{G} \sum_{i=1}^{G} A^i\\
    &\left[\frac{1}{K} \sum_{j=1}^{K} \left( \prod_{t \in ch_j} \frac{p_\theta \left( x_{t-1}^i | x_t^i, c \right)}{p_{\theta_\text{old}} \left( x_{t-1}^i | x_t^i, c \right)} \right) ^{\frac{1}{cs_j}} 
    \cdot \frac{1}{cs_j} \sum_{t \in ch_j}
    \nabla_\theta \text{log} \left( p_\theta (x^i_{t-1} | x^i_t, c) \right) \right]
\end{aligned}
\end{equation}
Therefore, the  fundamental distinction between chunk-level policy optimization and step-level GRPO lies in how they weight the gradients of the log likelihoods. In step-level GRPO, they are weighted individually according to their respective importance weight $\frac{p_\theta (x^i_{t-1} | x^i_t, c)}{p_\text{old} (x^i_{t-1} | x^i_t, c)}$. However, these unequal weights, which can vary among $(0,1 + \epsilon]$ for $A \geq 0$ or $[1 - \epsilon, \infty)$ for $A \leq 0$, are not negligible, and their impact can accumulate and lead to unstable consequences. In contrast, chunk-level policy optimization applies a unified weight $\left( \prod_{t \in ch_j} \frac{p_\theta \left( x_{t-1}^i | x_t^i, c \right)}{p_{\theta_\text{old}} \left( x_{t-1}^i | x_t^i, c \right)} \right) ^{\frac{1}{cs_j}}$ to all timesteps within a chunk, effectively smoothing these fluctuations and eliminating training instability.

\section{Details}

\subsection{Details for \Cref{figure_quality_main}} \label{app_fig_detail}

The ``GCPO w/o Temporal Dynamics" variant in \Cref{figure_quality_main} is with the setting that all steps are treated as a single chunk ($k=1$). 

\subsection{Details for the Frequency of Inaccurate Advantage Attribution} \label{Details for the Frequency of Inaccurate Advantage Attribution}

We used Flux \cite{flux2024} as the test model and followed its default parameters for generation. Specifically, the generation timestep is $T=50$. All final rewards and step advantages are obtained by the step-aware preference model from \cite{liang2025aesthetic}. 

\subsection{Details for the Process Reward Comparison} \label{Details for the Process Reward Comparison}

\begin{wraptable}{r}{0.3\textwidth}
\vskip -0.3in
\centering
\caption{Results of Process Reward}
\begin{tabular}{lc}
\toprule
\textbf{Method} & \textbf{HPSv2} \\
\midrule
Flux & 0.304 \\
\midrule
Dance-GRPO & \cellcolor{C1!50!white} 0.365 \\
Process Reward & 0.348 \\
\bottomrule
\end{tabular}
\vskip -0.2in
\label{table_process_reward}
\end{wraptable}

To validate the limitations of current process reward models, we replaced the reward in GRPO \cite{shao2024deepseekmath, guo2025deepseek} with the process reward from \cite{liang2025aesthetic}, and compared it to the standard GRPO. The evaluation metric is HPSv2 \cite{wu2023human}. The results in \Cref{table_process_reward} demonstrates the limitations of current process reward models.

\subsection{Details for Training} \label{Details for Training}

All experiments were conducted on $8$ Nvidia GPUs. The hyperparameters are summarized in \Cref{table_hyperparams}, remaining the same for all baselines and our approach.

\begin{table}[htbp]
\centering
\vskip -0.2in
\caption{Hyperparameter Settings}
\label{table_hyperparams}
\renewcommand{\arraystretch}{1.2}
\setlength{\tabcolsep}{8pt}
\begin{tabular}{l l l l}
\toprule
\textbf{Parameter} & \textbf{Value} & \textbf{Parameter} & \textbf{Value} \\
\midrule
Learning rate & $1\times 10^{-5}$ & Weight decay & $1\times 10^{-4}$ \\
Train batch size & 2 & SP size & 1 \\
SP batch size & 2 & Max grad norm & 0.01 \\
Resolution & $720\times720$ & Sampling steps & 17 \\
Eta & 0.7 & Num. generations & 12 \\
Grad. accum. steps & 12 & Shift (branch offset) & 3 \\
Clip range & $5\times 10^{-5}$ & Training steps & 150 \\
Timestep fraction & $0.5$ &  &  \\
\bottomrule
\end{tabular}
\vskip -0.1in
\end{table}

\subsection{Details for Evaluation} \label{Details for Evaluation}

\begin{wraptable}{r}{0.5\textwidth}
\vskip -0.15in
\centering
\caption{Performance on Preference Alignment.}
\begin{tabular}{lcc}
\toprule
\textbf{Methods} & \textbf{HPSv3} & \textbf{ImageReward} \\
\midrule
Flux & 0.304 & 1.086 \\
\midrule
Dance-GRPO & 15.080 & 1.141 \\
Flow-GRPO & 14.900 & 1.135 \\
Pref-GRPO & 14.868 & 1.139 \\
MixGRPO & 15.128 & 1.146 \\
TP-GRPO & 15.206 & 1.148 \\
GCPO & \cellcolor{C1!50!white} 15.236 & \cellcolor{C1!50!white} 1.149 \\
\bottomrule
\end{tabular}
\vskip -0.4in
\label{table_preference_alignment_app}
\end{wraptable}

We set the generation timestep $T=50$ during evaluation. Following \citep{li2025mixgrpo}, the first 30 steps are sampled with the trained model, while the remaining 20 steps are sampled with the base model. This hybrid inference strategy and corresponding settings, also used in \citep{li2025mixgrpo}, have proven effective in mitigating reward hacking.

\subsection{Details for User Study} \label{Details for User Study}

During the test phase, ten reviewers possessing good aesthetic tastes were recruited. We extracted 40 prompts from our evaluation dataset - HPDv2.1 test set\citep{wu2023human}.

\section{Additional Experiments}

\subsection{Additional Baselines} \label{additional_baselines_app}

We provide performance comparison with additional baselines: Pref-GRPO \citep{wang2025pref}, MixGRPO \citep{li2025mixgrpo}, and TP-GRPO \citep{tong2026alleviating}, where TP-GRPO is a process-reward approach. Note that the weighted sampling strategy is disabled here. \Cref{table_preference_alignment_app} demonstrates that, our method outperforms the step-level approach by solving the sparse reward issue from the unified chunk-level importance ratio perspective. 

\subsection{Additional Ablation Studies} \label{additional_ablation_study_app}

\begin{wraptable}{r}{0.5\textwidth}
\vskip -0.2in
\centering
\caption{Ablation Results of Different Distance Metrix.}
\begin{tabular}{lcc}
\toprule
\textbf{Methods} & \textbf{HPSv3} & \textbf{ImageReward} \\
\midrule
Flux & 0.304 & 1.086 \\
\midrule
$L1$ & 15.236 & 1.147 \\
$L2$ & \cellcolor{C1!50!white} 15.268 & \cellcolor{C1!50!white} 1.149 \\
Cosine & 15.220 & 1.147 \\
\bottomrule
\end{tabular}
\vskip -0.4in
\label{table_ablation_app}
\end{wraptable}

We used relative $L1$ distance $L1_\text{rel} (x, t)$ to capture temporal dynamics in the implementation. However, there are also other options. Additionally, we conduct ablation studies comparing $L1$, $L2$, and cosine similarity. \Cref{table_ablation_app} shows that despite $L2$ even achieves better performance in some cases, the consistent improvement over various distance matrices highlights the effectiveness of chunk-level optimization.

\section{Additional Visualization} 

\subsection{Main Visualization} \label{Main Visualization}

\Cref{Additional Qualitative Results}, \Cref{Additional Qualitative Results2}, \Cref{Additional Qualitative Results3}, and \Cref{Additional Qualitative Results4} present qualitative comparisons among FLUX, Dance-GRPO, GCPO without temporal dynamics, and GCPO with temporal dynamics. Overall, GCPO generates outputs that align more closely with human aesthetics, exhibiting stronger lighting contrast, more vivid colors, and finer details. Notably, the contrast between columns $3$ and $4$ in \Cref{Additional Qualitative Results} and \Cref{Additional Qualitative Results2} underscores the importance of chunking with the inherent temporal dynamics of flow matching, which substantially elevates aesthetic style and structure of images.

\begin{figure}[h]
\centering
\includegraphics[width=0.8\textwidth]{figures/quality_app1.jpg}
\caption{Additional visualization comparison between the FLUX, DanceGRPO, and GCPO.}
\label{Additional Qualitative Results}
\end{figure}

\begin{figure}[h]
\centering
\includegraphics[width=0.8\textwidth]{figures/quality_app2.jpg}
\caption{Additional visualization comparison between the FLUX, DanceGRPO, and GCPO.}
\label{Additional Qualitative Results2}
\end{figure}

\begin{figure}[h]
\centering
\includegraphics[width=0.9\textwidth]{figures/quality_app3.jpg}
\caption{Additional visualization comparison between the FLUX, DanceGRPO, and GCPO. The GCPO here is with temporal dynamics.}
\label{Additional Qualitative Results3}
\end{figure}

\begin{figure}[h]
\centering
\includegraphics[width=0.9\textwidth]{figures/quality_app4.jpg}
\caption{Additional visualization comparison between the FLUX, DanceGRPO, and GCPO. The GCPO here is with temporal dynamics.}
\label{Additional Qualitative Results4}
\end{figure}

\subsection{Visualization Comparison of the Weighted Sampling Strategy} \label{Visualization Comparison of the Weighted Sampling Strategy}

\Cref{Visualization Comparison} presents additional qualitative comparisons of the weighted sampling strategy (the last column). While this strategy improves preference alignment, it destabilizes image structures in high-noise regions.

\begin{figure}[h]
\centering
\includegraphics[width=1.0\textwidth]{figures/com_app.jpg}
\caption{Additional visualization comparison of the Weighted Sampling Strategy (the last column).}
\label{Visualization Comparison}
\end{figure}

%%%%%%%%%%%%%%%%%%%%%%%%%%%%%%%%%%%%%%%%%%%%%%%%%%%%%%%%%%%%%%%%%%%%%%%%%%%%%%%
%%%%%%%%%%%%%%%%%%%%%%%%%%%%%%%%%%%%%%%%%%%%%%%%%%%%%%%%%%%%%%%%%%%%%%%%%%%%%%%

\end{document}